\documentclass[11pt,a4paper]{article}
\usepackage[hyperref]{acl2020}
\usepackage{times}
\usepackage{latexsym}

\usepackage{graphicx}
\usepackage{float}
\usepackage{multirow}
\usepackage{subfigure}

\usepackage{microtype}

\aclfinalcopy 

\title{Applying Cyclical Learning Rate to Neural Machine Translation}

\author{Choon Meng Lee, Jianfeng Liu, Wei Peng \thanks{ Corresponding author}\\
Artificial Intelligence Application Research Center, Huawei Technologies \\
\texttt{\{lee.choonmeng,liujianfeng,peng.wei1\}@huawei.com} \\
}

\date{}

\begin{document}
\maketitle
\begin{abstract}
In training deep learning networks, the optimizer and related learning rate are often used without much thought or with minimal tuning, even though it is crucial in ensuring a fast convergence to a good quality minimum of the loss function that can also generalize well on the test dataset. Drawing inspiration from the successful application of cyclical learning rate policy for computer vision related convolutional networks and datasets, we explore how cyclical learning rate can be applied to train transformer-based neural networks for neural machine translation. From our carefully designed experiments, we show that the choice of optimizers and the associated cyclical learning rate policy can have a significant impact on the performance. In addition, we establish guidelines when applying cyclical learning rates to neural machine translation tasks. Thus with our work, we hope to raise awareness of the importance of selecting the right optimizers and the accompanying learning rate policy, at the same time, encourage further research into easy-to-use learning rate policies.

\end{abstract}

\section{Introduction}
\label{sec:introduction}
There has been many interests in deep learning optimizer research recently \citep{amsgrad, adabound, lookahead_optimizer, radam}. These works attempt to answer the question: what is the best step size to use in each step of the gradient descent? With the first order gradient descent being the \textit{de facto} standard in deep learning optimization, the question of the optimal step size or learning rate in each step of the gradient descent arises naturally. The difficulty in choosing a good learning rate can be better understood by considering the two extremes: 1) when the learning rate is too small, training takes a long time; 2) while overly large learning rate causes training to diverge instead of converging to a satisfactory solution. 

The two main classes of optimizers commonly used in deep learning are the momentum based Stochastic Gradient Descent (SGD) \citep{sgd} and adaptive momentum based methods \citep{adagrad, adam, amsgrad, adabound, radam}. The difference between the two lies in how the newly computed gradient is updated. In SGD with momentum, the new gradient is updated as a convex combination of the current gradient and the exponentially averaged previous gradients. For the adaptive case, the current gradient is further weighted by a term involving the sum of squares of the previous gradients. For a more detailed description and convergence analysis, please refer to \citet{amsgrad}.

In Adam \citep{adam}, the experiments conducted on the MNIST and CIFAR10 dataset showed that Adam has the fastest convergence property, compared to other optimizers, in particular SGD with Nesterov momentum. Adam has been popular with the deep learning community due to the speed of convergence. However, Adabound \citep{adabound}, a proposed improvement to Adam by clipping the gradient range, showed in the experiments that given enough training epochs, SGD can converge to a better quality solution than Adam. To quote from the future work of Adabound, ``why SGD usually performs well across diverse applications of machine learning remains uncertain". The choice of optimizers is by no means straight forward or cut and dry.

Another critical aspect of training a deep learning model is the batch size. Once again, while the batch size was previously regarded as a hyperparameter, recent studies such as \citet{keskar_sharp} have shed light on the role of batch size when it comes to generalization, i.e., how the trained model performs on the test dataset. Research works \citep{keskar_sharp, flat_minima} expounded the idea of sharp vs. flat minima when it comes to generalization. From experimental results on convolutional networks, e.g., AlexNet \citep{alexnet}, VggNet \citep{vgg},  \citet{keskar_sharp} demonstrated that overly large batch size tends to lead to sharp minima while sufficiently small batch size brings about flat minima. \citet{sharp_minima_generalization}, however, argues that sharp minima can also generalize well in deep networks, provided that the notion of sharpness is taken in context.

While the aforementioned works have helped to contribute our understanding of the nature of the various optimizers, their learning rates and batch size effects, they are mainly focused on computer vision (CV) related deep learning networks and datasets. In contrast, the rich body of works in Neural Machine Translation (NMT) and other Natural Language Processing (NLP) related tasks have been largely left untouched. Recall that CV deep learning networks and NMT deep learning networks are very different. For instance, the convolutional network that forms the basis of many successful CV deep learning networks is translation invariant, e.g., in a face recognition network, the convolutional filters produce the same response even when the same face is shifted or translated. In contrast, Recurrent Neural Networks (RNN) \citep{lstm, gru} and transformer-based deep learning networks \citep{Vaswani2017, bert} for NMT are specifically looking patterns in sequences. There is no guarantee that the results from the CV based studies can be carried across to NMT. There is also a lack of awareness in the NMT community when it comes to optimizers and other related issues such as learning rate policy and batch size. It is often assumed that using the mainstream optimizer (Adam) with the default settings is good enough. As our study shows, there is significant room for improvement.

\subsection{The Contributions}
The contributions of this study are to:
\begin{itemize}
  \item Raise awareness of how a judicial choice of optimizer with a good learning rate policy can help improve performance;
  \item Explore the use of cyclical learning rates for NMT. As far as we know, this is the first time cyclical learning rate policy has been applied to NMT;
  \item Provide guidance on how cyclical learning rate policy can be used for NMT to improve performance. 
\end{itemize}

\section{Related Works}
\citet{viz_loss} proposes various visualization methods for understanding the loss landscape defined by the loss functions and how the various deep learning architectures affect the landscape. The proposed visualization techniques allow a depiction of the optimization trajectory, which is particularly helpful in understanding the behavior of the various optimizers and how they eventually reach their local minima.

Cyclical Learning Rate (CLR) \citep{clr} addresses the learning rate issue by having repeated cycles of linearly increasing and decreasing learning rates, constituting the triangle policy for each cycle. CLR draws its inspiration from curriculum learning \citep{curriculum} and simulated annealing \citep{sim_anneal}. \citet{clr} demonstrated the effectiveness of CLR on standard computer vision (CV) datasets CIFAR-10 and CIFAR-100, using well established CV architecture such as ResNet \citep{resnet} and DenseNet \citep{densenet}. As far as we know, CLR has not been applied to Neural Machine Translation (NMT). The methodology, best practices and experiments are mainly based on results from CV architecture and datasets. It is by no means apparent or straightforward that the same approach can be directly carried over to NMT.

One interesting aspect of CLR is the need to balance regularizations such as weight decay, dropout and batch size, etc., as pointed out in  \citet{super_clr}. The experiments verified that various regularizations need to be toned down when using CLR to achieve good results. In particular, the generalization results using the small batch size from the above-mentioned studies no longer hold for CLR. This is interesting because the use of CLR allows training to be accelerated by using a larger batch size without the sharp minima generalization concern. A related work is \citet{empirical_batchsize}, which sets a theoretical upper limit on the speed up in training time with increasing batch size. Beyond this theoretical upper limit, there will be no speed up in training time even with increased batch size.

\section{The Proposed Approach}
\label{sec:methods}
\begin{figure}[!h]
    \centering
    \includegraphics[width=0.47\textwidth]{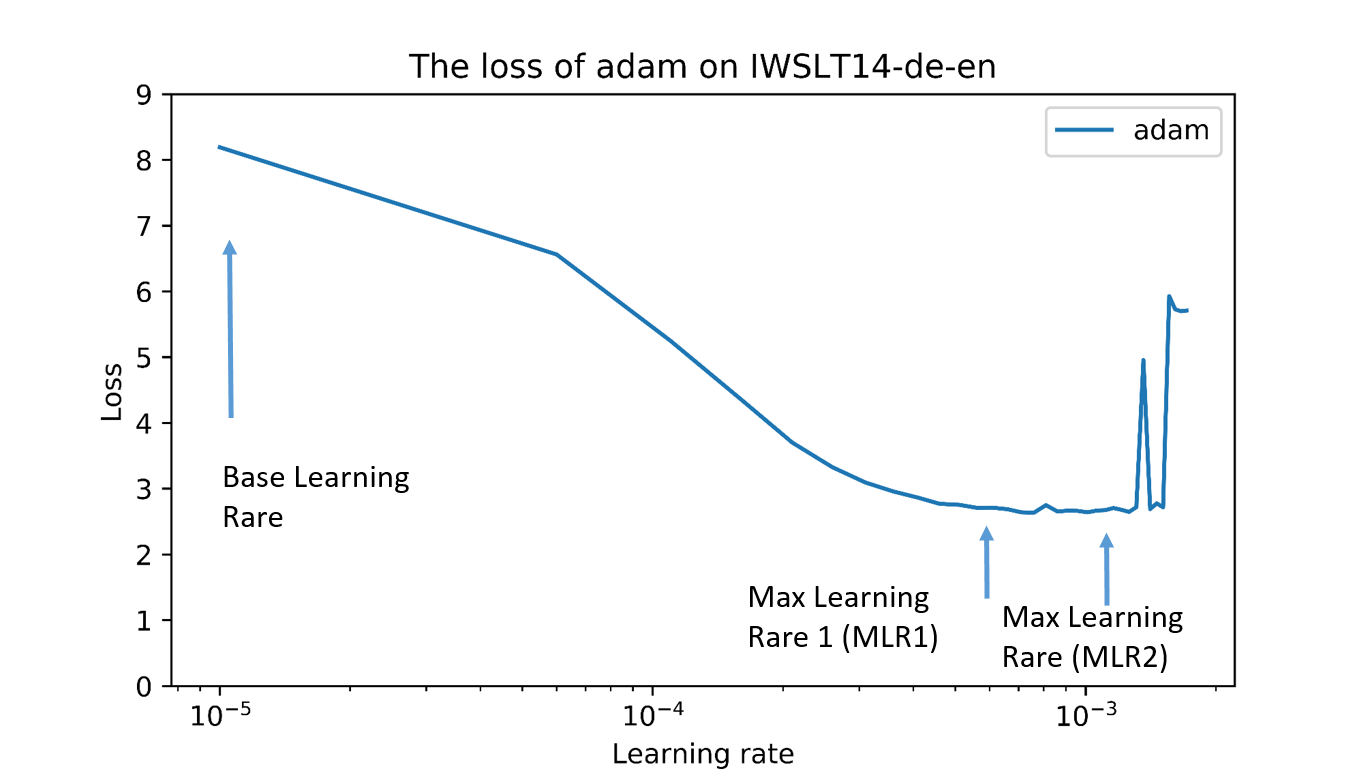}
    \caption{Range test curve for the IWSLT2014-de-en dataset, showing the chosen base and max learning rate for the triangular policy.}
    \label{range_test_IWSLT14-de-en}
\end{figure}

Our main approach in the NMT-based learning rate policy is based on the triangular learning rate policy in CLR. For CLR, some pertinent parameters need to be determined: base/max learning rate and cycle length. As suggested in CLR, we perform the range test to set the base/max learning rate while the cycle length is some multiples of the number of epochs. The range test is designed to select the base/max learning rate in CLR. Without the range test, the base/max learning rate in CLR will need to be tuned as hyperparameters which is difficult and time consuming. In a range test, the network is trained for several epochs with the learning rate linearly increased from an initial rate. For instance, the range test for the IWSLT2014 (DE2EN) dataset was run for 35 epochs,  with the initial learning rate set to some small values, e.g., $1 \times 10^{-5}$ for Adam and increased linearly over the 35 epochs. Given the range test curve, e.g., Figure \ref{range_test_IWSLT14-de-en}, the base learning rate is set to the point where the loss starts to decrease while the maximum learning rate is selected as the point where the loss starts to plateau or to increase. As shown in Figure \ref{range_test_IWSLT14-de-en}, the base learning rate is selected as the initial learning rate for the range test, since there is a steep loss using the initial learning rate. The max learning rate is the point where the loss stagnates. For the step size, following the guideline given in \citet{clr} to select the step size between 2-10 times the number of iterations in an epoch and set the step size to 4.5 epochs.

\begin{figure}[!h]
    \centering
    \includegraphics[width=0.47\textwidth]{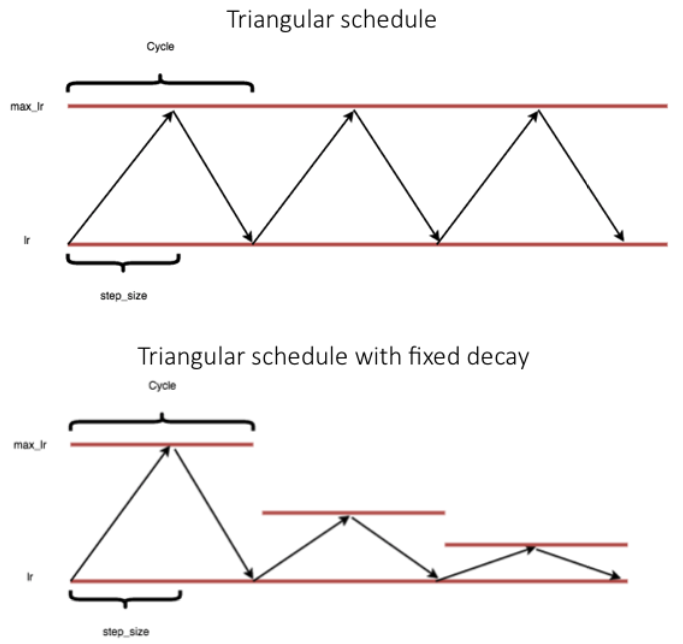}
    \caption{The learning rate decay used in our experiments.}
    \label{clr_decay}
\end{figure}

The other hyperparameter to take care of is the learning rate decay rate, shown in Figure \ref{clr_decay}. For the various optimizers, the learning rate is usually decayed to a small value to ensure convergence. There are various commonly used decay schemes such as piece-wise constant step function, inverse (reciprocal) square root. This study adopts two learning rate decay policies: 
\begin{itemize}
  \item Fixed decay (shrinking) policy where the max learning rate is halved after each learning rate cycle;
  \item No decay. This is unusual because for both SGD and adaptive momentum optimizers, a decay policy is required to ensure convergence.
\end{itemize}
Our adopted learning rate decay policy is interesting because experiments in \citet{clr} showed that using a decay rate is detrimental to the resultant accuracy. Our designed experiments in Section \ref{sec:experiment} reveal how CLR performs with the chosen decay policy.

\begin{table*}[h]
\centering
\begin{tabular}{lccccc}
\hline \textbf{Corpus} & \textbf{Train} & \textbf{Valid.} &\textbf{Test} & \textbf{Source Vocab.} & \textbf{Target Vocab.} \\ \hline
IWSLT2014-de-en (DE2EN) & 160,239 & 7,283 & 6,750 & 8,844 & 6,628 \\
IWSLT2014-fr-en (FR2EN) & 166,045 & 4,818 & 4,800 & 8,508 & 7,308 \\
IWSLT2017-de-en (DE2EN) & 192,347 & 4,829 & 4,822 & 13,156 & 10,108 \\
\hline
\end{tabular}
\caption{\label{data-table} Datasets used for the experiment. }
\end{table*}

The CLR decay policy should be contrasted with the standard inverse square root policy (INV) that is commonly used in deep learning platforms, e.g., in fairseq \citep{ott2019fairseq}. The inverse square root policy (INV) typically starts with a warm-up phase where the learning rate is linearly increased to a maximum value. The learning rate is decayed as the reciprocal of the square root of the number of epochs from the above-mentioned maximum value. 

The other point of interest is how to deal with batch size when using CLR. Our primary interest is to use a larger batch size without compromising the generalization capability on the test set. Following the lead in \citet{super_clr}, we look at how the NMT tasks perform when varying the batch size on top of the CLR policy. Compared to \citet{super_clr}, we stretch the batch size range, going from batch size as small as 256 to as high as 4,096. Only through examining the extreme behaviors can we better understand the effect of batch size superimposed on CLR.

\section{Experiments}
\label{sec:experiment}

\subsection{Experiment Settings}
The purpose of this section is to demonstrate the effects of applying CLR and various batch sizes to train NMT models. The experiments are performed on two translation directions (DE $\rightarrow$ EN and FR $\rightarrow$ EN) for IWSLT2014 and IWSLT2017 \citep{cettoloEtAl:EAMT2012}.    

The data are pre-processed using functions from Moses \citep{moses}.  The punctuation is normalized into a standard format. After tokenization, byte pair encoding (BPE) \citep{sennrich2016b} is applied to the data to mitigate the adverse effects of out-of-vocabulary (OOV) rare words. The sentences with a source-target sentence length ratio greater than 1.5 are removed to reduce potential errors from sentence misalignment. Long sentences with a length greater than 250 are also removed as a common practice. The split of the datasets produces the training, validation (valid.) and test sets presented in Table \ref{data-table}. 

\begin{table}[h]
\begin{center}
\begin{tabular}{ll}
\hline \textbf{Hyperparameters} & \textbf{Values} \\ \hline
Encoder/Decoder Layers & 6  \\
Embedding Units & 512 \\
Attention Heads & 4 \\
Feed-forward Hidden Units & 1,024 \\
Batch Size (default) & 4,096 \\
Training Epoch (default) & 50 \\
\hline
\end{tabular}
\end{center}
\caption{\label{para-table} Hyperparameters for the experiments.}
\end{table}

The transformer architecture \citep{Vaswani2017} from fairseq \citep{ott2019fairseq} \footnote{https://github.com/pytorch/fairseq} is used for all the experiments. The hyperparameters are presented in Table \ref{para-table}. We compared training under CLR with an inverse square for two popular optimizers used in machine translation tasks, Adam and SGD. All models are trained using one NVIDIA V100 GPU.

\begin{table*}[h]
\centering
\begin{tabular}{lcccc}
\hline
\multirow{2}{*}{\textbf{Corpus}} & \multicolumn{2}{c}{\textbf{Adam}} & \multicolumn{2}{c}{\textbf{SGD}} \\ \cline{2-5} 
 & \textbf{Max} & \textbf{Base} & \textbf{Max} & \textbf{Base} \\ \hline
IWSLT2014-de-en & 5.00E-04 & 1.00E-05 & 6.90E+00 & 1.00E-03 \\
IWSLT2014-fr-en & 8.00E-04 & 1.00E-05 & - & - \\
IWSLT2017-de-en & 7.60E-04 & 1.00E-05 & 8.00E+00 & 1.00E-03 \\
\hline
\end{tabular}
\caption{\label{lrb-table} Learning rate boundary for CLR. }
\end{table*}



The learning rate boundary of the CLR is selected by the range test (shown in Figure \ref{range_test_IWSLT14-de-en}). The base and maximal learning rates adopted in this study are presented in Table \ref{lrb-table}. 
Shrink strategy is applied when examining the effects of CLR in training NMT. The optimizers (Adam and SGD) are assigned with two options: 1) without shrink (as ``nshrink"); 2) with shrink at a rate of 0.5 (``yshrink"), which means the maximal learning rate for each cycle is reduced at a decay rate of 0.5.  

\subsection{Effects of Applying CLR to NMT Training}
\begin{figure}[h]
    \centering
    \includegraphics[width=0.47\textwidth]{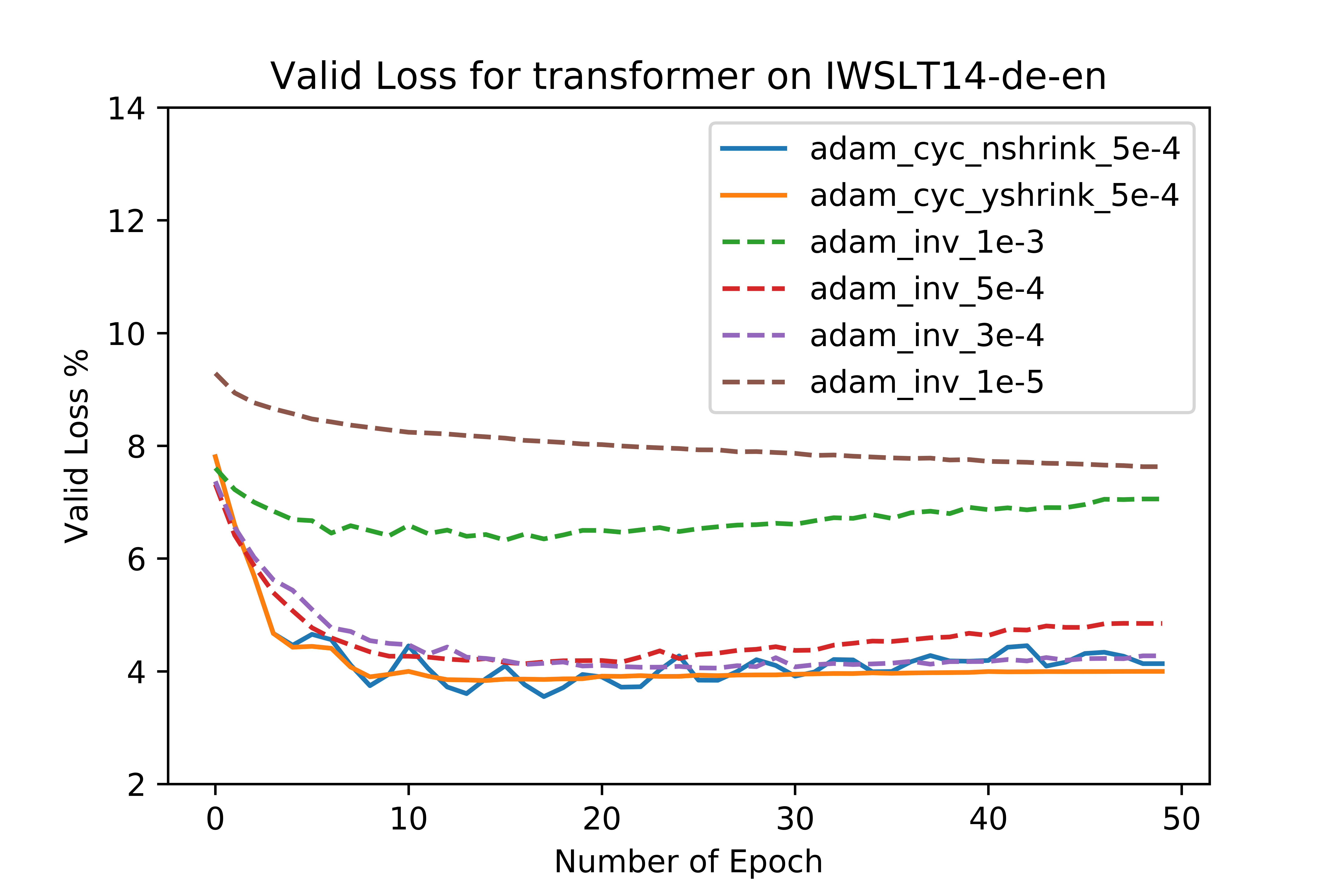}
    \caption{A comparison study of training NMT models on IWSLT2014-de-en using CLR and INV with a range of initial learning rate on Adam. The learning rate policy ``adam\_cyc\_nshink\_5e-4" denotes the optimizer Adam is trained under CLR with the no shrink option and a max learning rate of 5e-4. }
    \label{fig3}
\end{figure}

\begin{figure}[h]
    \centering
    \includegraphics[width=0.47\textwidth]{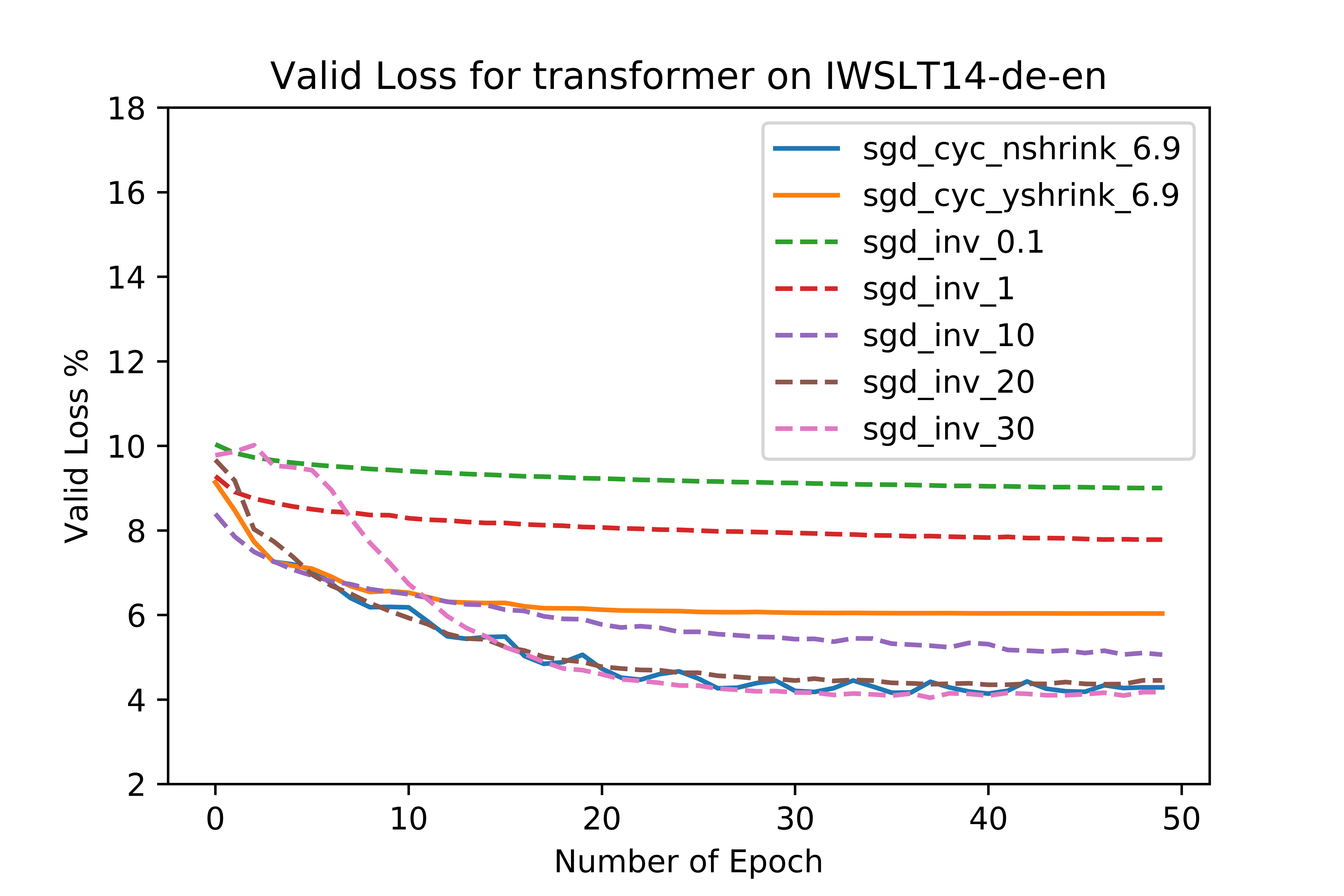}
    \caption{A comparison study of training NMT models on IWSLT2014-de-en using CLR and INV with a range of initial learning rate on SGD. The learning rate policy ``sgd\_cyc\_yshink\_5e-4" denotes the optimizer SGD is trained under CLR with the shrink option and a max learning rate of 5e-4.}
    \label{fig4}
\end{figure}

A hypothesis we hold is that NMT training under CLR may result in a better local minimum than that achieved by training with the default learning rate schedule. A comparison experiment is performed for training NMT models for ``IWSLT2014-de-en" corpus using CLR and INV with a range of initial learning rates on two optimizers (Adam and SGD), respectively. 
It can be observed that both Adam and SGD are very sensitive to the initial learning rate under the default INV schedule before CLR is applied (as shown in Figures \ref{fig3} and \ref{fig4}). In general, SGD prefers a bigger initial learning rate when CLR is not applied. The initial learning rate of Adam is more concentrated towards the central range.

\begin{figure}[h]
    \centering
    \includegraphics[width=0.47\textwidth]{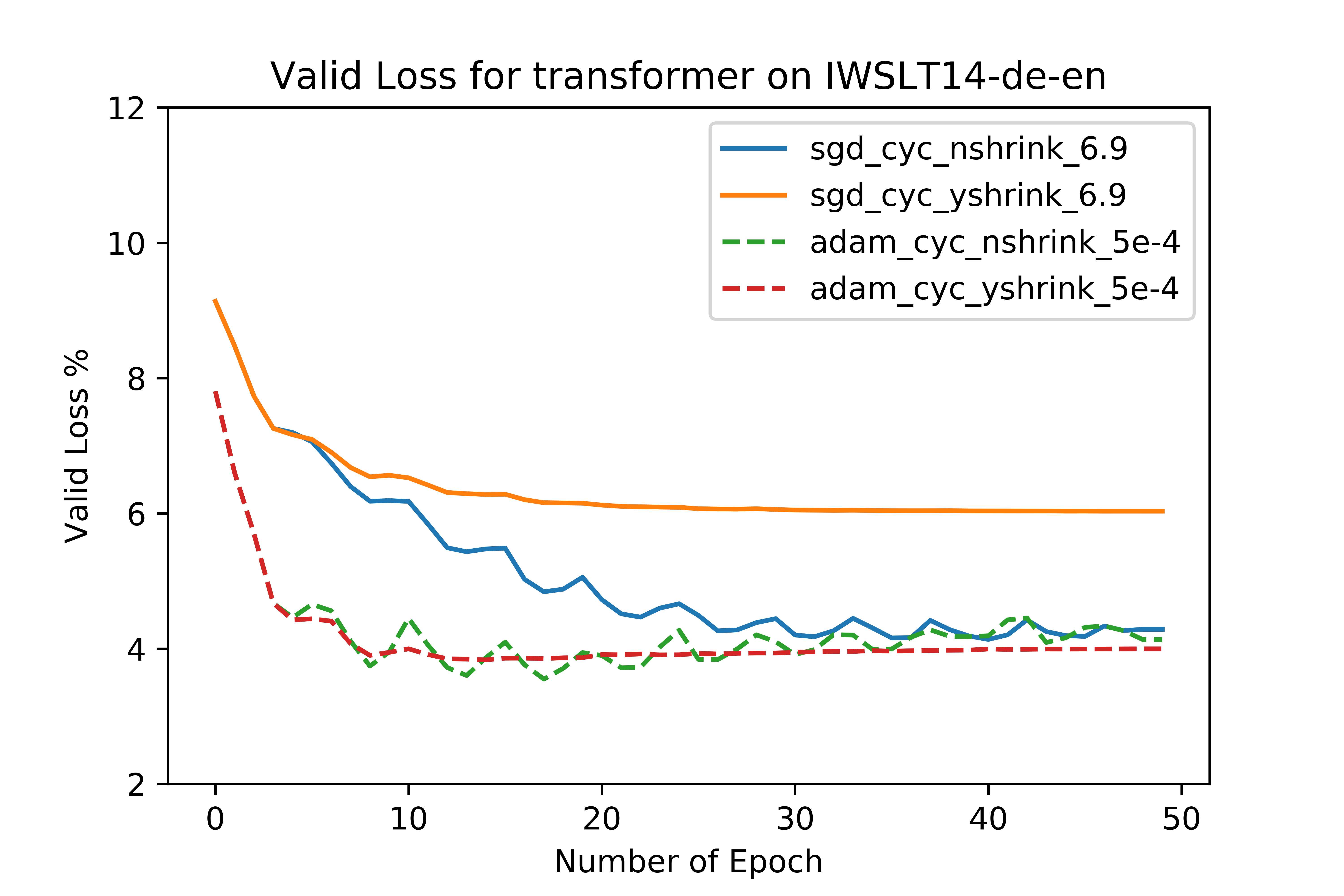}
    \caption{A view of effects of applying CLR to Adam and SGD when training the NMT on IWSLT2014-de-en.}
    \label{fig5}
\end{figure}

Applying CLR has positive impacts on NMT training for both Adam and SGD. When applied to SGD, CLR exempts the needs for a big initial learning rate as it enables the optimizer to explore the local minima better.  Shrinking on CLR for SGD is not desirable as a higher learning rate is required (Figure \ref{fig4}). It is noted that applying CLR to Adam produces consistent improvements regardless of shrink options (Figure \ref{fig3}). Furthermore, it can be observed that the effects of applying CLR to Adam are more significant than those of SGD, as shown in Figure \ref{fig5}. Similar results are obtained from our experiments on ``IWSLT2017-de-en" and ``IWSLT2014-fr-en" corpora (Figures \ref{figapendix1} and \ref{figappendix2} in Appendix~\ref{sec:appendix}). The corresponding BLEU scores are presented in Table \ref{bleu-table}, in which the above-mentioned effects of CLR on Adam can also be established. The training takes fewer epochs to converge to reach a local minimum with better BLEU scores (i.e., bold fonts in Table \ref{bleu-table}).               

\begin{table*}[h]
\centering
\begin{tabular}{lccc}
\hline
\textbf{Corpus} & \textbf{Learning Rate Policy} & \textbf{Best BLEU} & \textbf{Epoch} \\ \hline
\multirow{5}{*}{IWSLT2014-de-en} & adam\_cyc\_nshrink\_5e-4 & \textbf{32.65} & \textbf{18} \\
 & adam\_cyc\_yshrink\_5e-4 & \textbf{31.29} & \textbf{18} \\
 & adam\_inv\_5e-4 & 30.88 & 16 \\
 & sgd\_inv\_30 & 30.78 & 42 \\
 & adam\_inv\_3e-4 & 30.46 & 34 \\
 & sgd\_cyc\_nshrink\_6.9 & 30.16 & 45\\
 \hline
\multirow{5}{*}{IWSLT2017-de-en} & adam\_cyc\_nshrink\_7.6e-4 & \textbf{33.00} & \textbf{18} \\
 & adam\_cyc\_yshrink\_7.6e-4 & \textbf{31.56} & \textbf{19} \\
 & sgd\_inv\_30 & 30.82 & 49 \\
 & adam\_inv\_3e-4 & 30.78 & 35 \\
 & adam\_inv\_5e-4 & 30.70 & 19 \\
 & sgd\_cyc\_nshrink\_8 & 30.40 & 49 \\
 & adam\_inv\_7.6e-4 & 28.94 & 40 \\
 \hline
\multirow{4}{*}{IWSLT2014-fr-en} & adam\_cyc\_nshrink\_8e-4 & \textbf{37.82} & \textbf{17} \\
 & adam\_cyc\_yshrink\_8e-4 & \textbf{36.91} & \textbf{17} \\
 & adam\_inv\_5e-4 & 36.43 & 17 \\
 & adam\_inv\_3e-4 & 36.25 & 35 \\
 & sgd\_inv\_30 & 35.51 & 45 \\ 
 & adam\_inv\_8e-4 & 6.20 & 43 \\
 \hline 
\end{tabular}
\caption{\label{bleu-table} The best BLEU for various learning rate policies when training NMT models on IWSLT2014-de-en, IWSLT2017-de-e and IWSLT2014-fr-en. The total number of training epochs for all the experiments is 50. The table is sorted by the best BLEU in descending order. }
\end{table*}






\subsection{Effects of Batch Size on CLR}
\begin{figure}[h]
    \centering
    \includegraphics[width=0.47\textwidth]{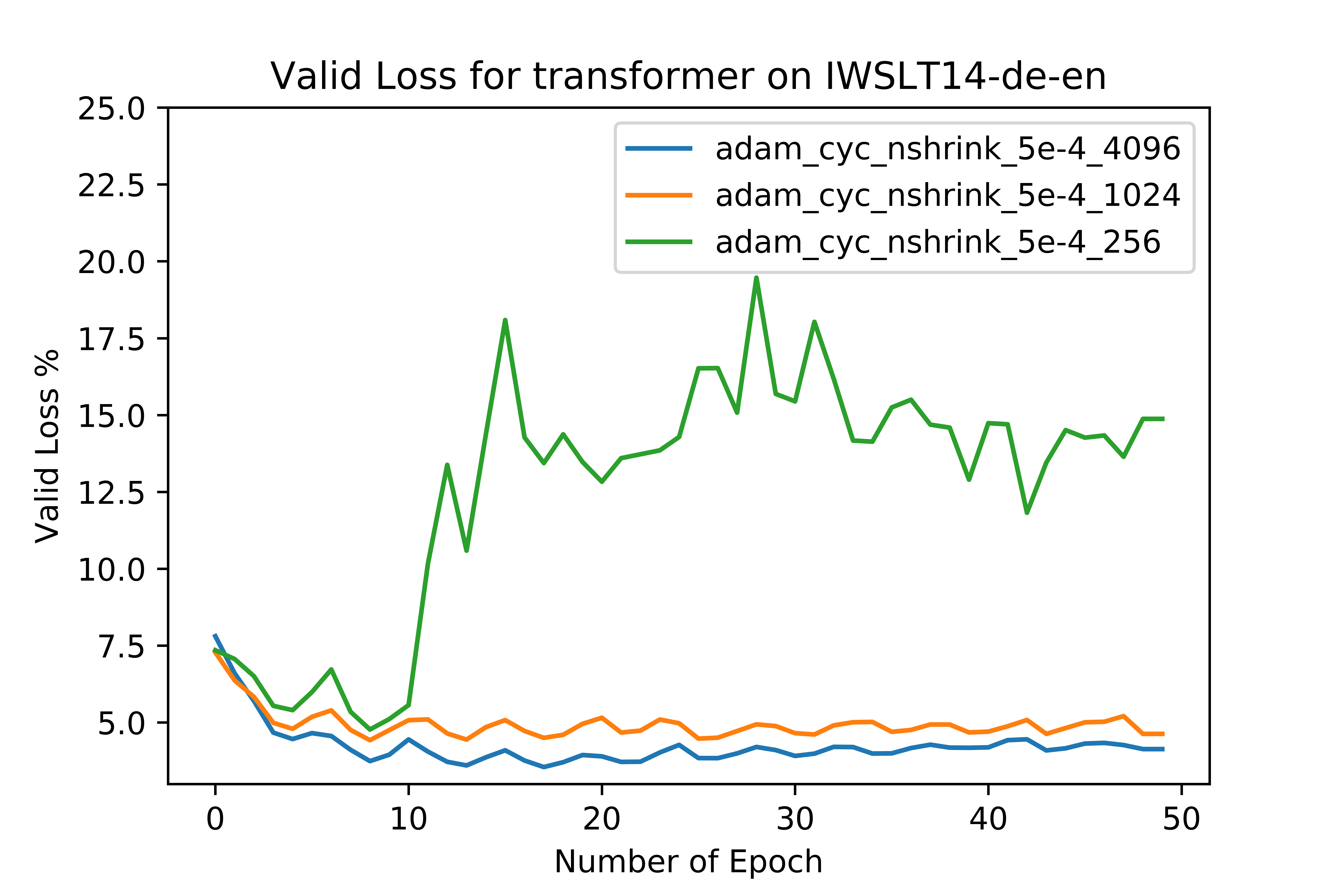}
    \caption{Effects of various batch sizes when training the NMT on IWSLT2014-de-en corpus with CLR.}
    \label{fig6}
\end{figure}

Batch size is regarded as a significant factor influencing deep learning models from the various CV studies detailed in Section \ref{sec:introduction}. It is well known to CV researchers that a large batch size is often associated with a poor test accuracy. However, the trend is reversed when the CLR policy is introduced by \citet{super_clr}. The critical question is: does this trend of using larger batch size with CLR hold for training transformers in NMT? Furthermore, what range of batch size does the associated regularization becomes significant? This will have implications because if CLR allows using a larger batch size without compromising the generalization capability, then it will allow training speed up by using a larger batch size. From Figure \ref{fig6}, we see that the trend of CLR with a larger batch size for NMT training does indeed lead to better performance. Thus the phenomenon we observe in \citet{super_clr} for CV tasks can be carried across to NMT. In fact, using a small batch size of 256 (the green curve in Figure \ref{fig6}) leads to divergence, as shown by the validation loss spiraling out of control. This is in line with the need to prevent over regularization when using CLR; in this case, the small batch size of 256 adds a strong regularization effect and thus need to be avoided. This larger batch size effect afforded by CLR is certainly good news because NMT typically deals with large networks and huge datasets. The benefit of a larger batch size afforded by CLR means that training time can be cut down considerably.


\label{sec:experiments}

\section{Further Analysis}
\begin{figure}[h]
    \centering
    \includegraphics[width=0.47\textwidth]{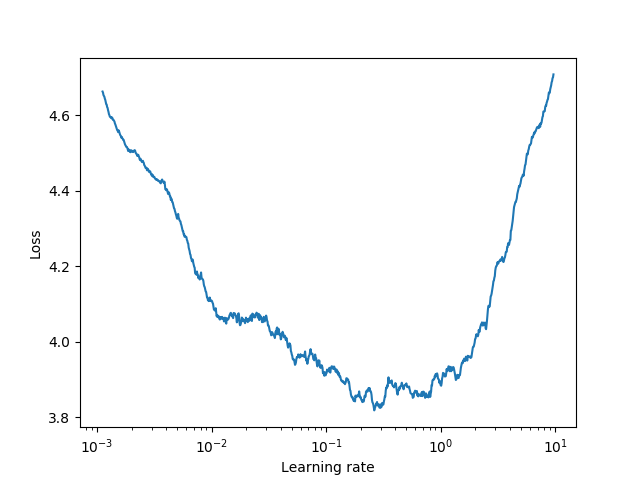}
    \caption{Range test curve for the CV CIFAR-100 dataset.}
    \label{range_test_CV}
\end{figure}

We observe the qualitative different range test curves for CV and NMT datasets. As we can see from Figures \ref{range_test_IWSLT14-de-en} and \ref{range_test_CV}. The CV range test curve looks more well defined in terms of choosing the max learning rate from the point where the curve starts to be ragged. For NMT, the range curve exhibits a smoother, more plateau characteristic. From Figure \ref{range_test_IWSLT14-de-en}, one may be tempted to exploit the plateau characteristic by choosing a larger learning rate on the extreme right end (before divergence occurs) as the triangular policy's max learning rate. From our experiments and empirical observations, this often leads to the loss not converging due to excessive learning rate. It is better to be more conservative and choose the point where the loss stagnates as the max learning rate for the triangular policy.
\subsection{How to Apply CLR to NMT Training Matters}
\begin{figure}[h]
    \centering
    \includegraphics[width=0.47\textwidth]{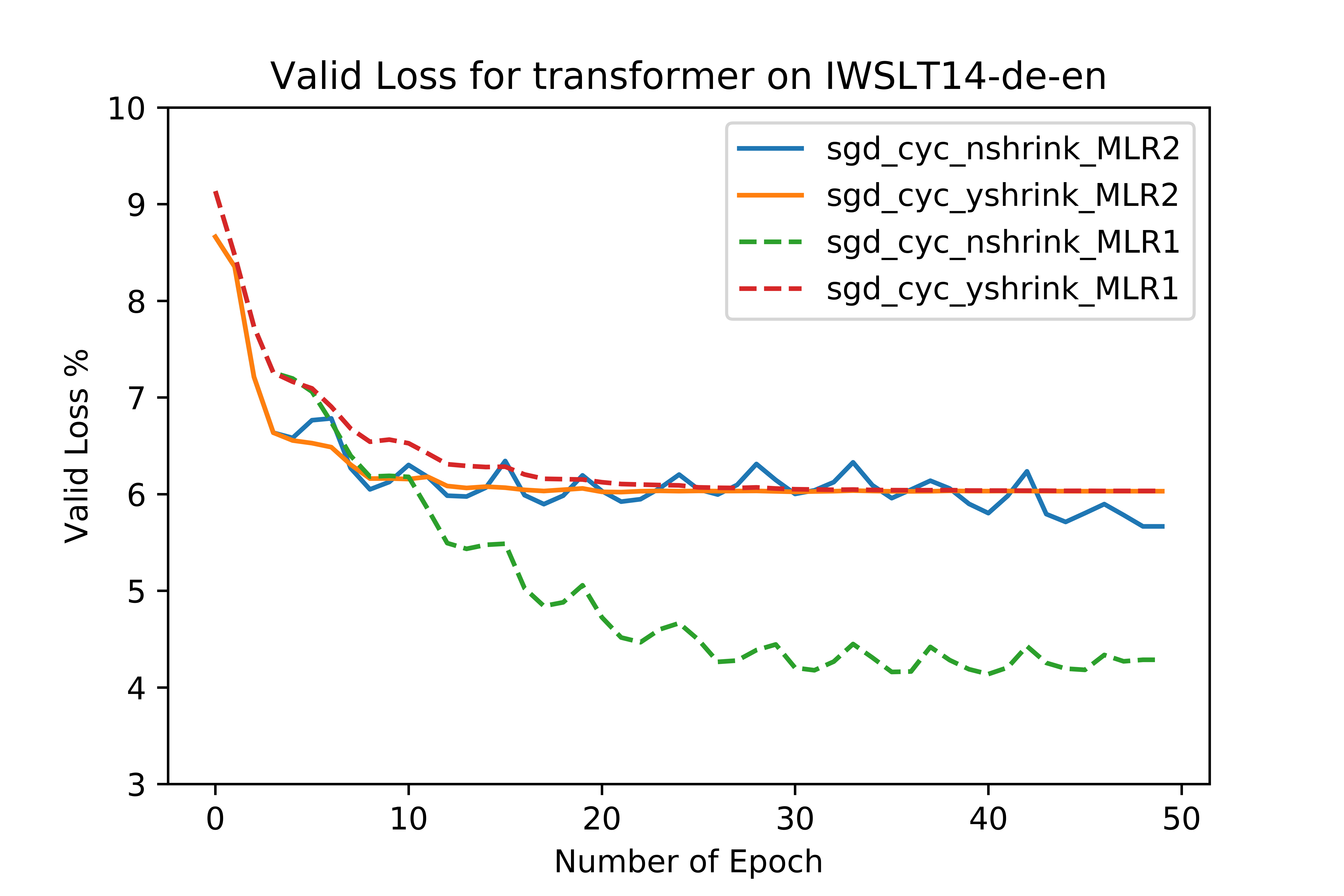}
    \caption{MLR1 with ``noshrink" is a preferable option for SGD when applying CLR to train NMT models on IWSLT2014-de-en.}
    \label{figureMLR1}
\end{figure}

\begin{figure}[h]
    \centering
    \includegraphics[width=0.47\textwidth]{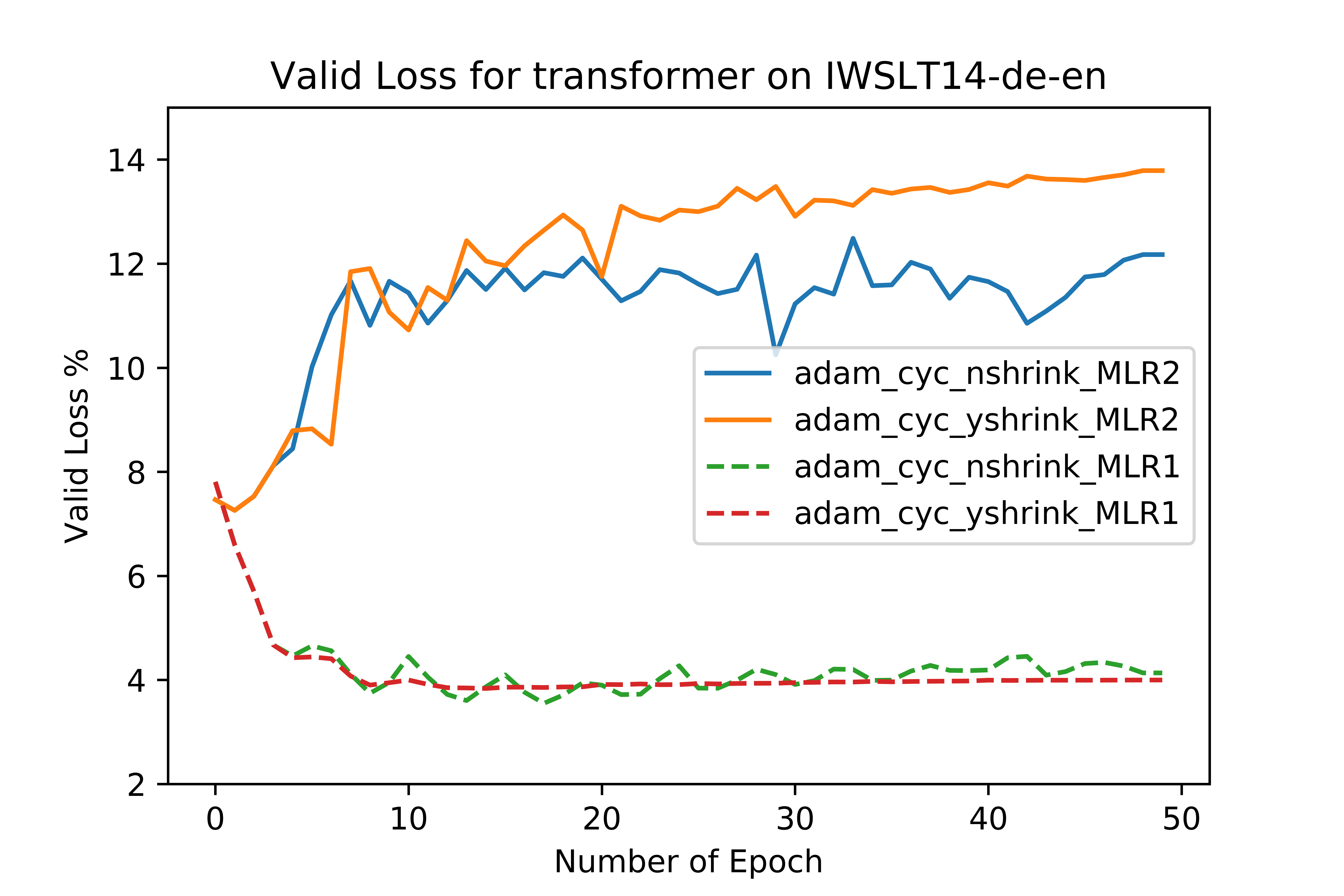}
    \caption{MLR1 is a preferable option for Adam when applying CLR to train NMT models on IWSLT2014-de-en.}
    \label{figureMLR2}
\end{figure}

A range test is performed to identify the max learning rates (MLR1 and MLR2) for the triangular policy of CLR (Figure \ref{range_test_IWSLT14-de-en}). The experiments showed the training is sensitive to the selection of MLR. As the range curve for training NMT models is distinctive to that obtained from a typical case of computer vision, it is not clear how to choose the MLR when applying CLR. A comparison experiment is performed to try MLRs with different values. It can be observed that MLR1 is a preferable option for both SGD and Adam (Figures \ref{figureMLR1} and \ref{figureMLR2}). The ``noshrink" option is mandatory for SGD, but this constraint can be relaxed for Adam. Adam is sensitive to excessive learning rate (MLR2).     

\begin{figure*}[h]
\centering                                                          
\subfigure[adam-inv-1e-3]{\includegraphics[scale=0.35]{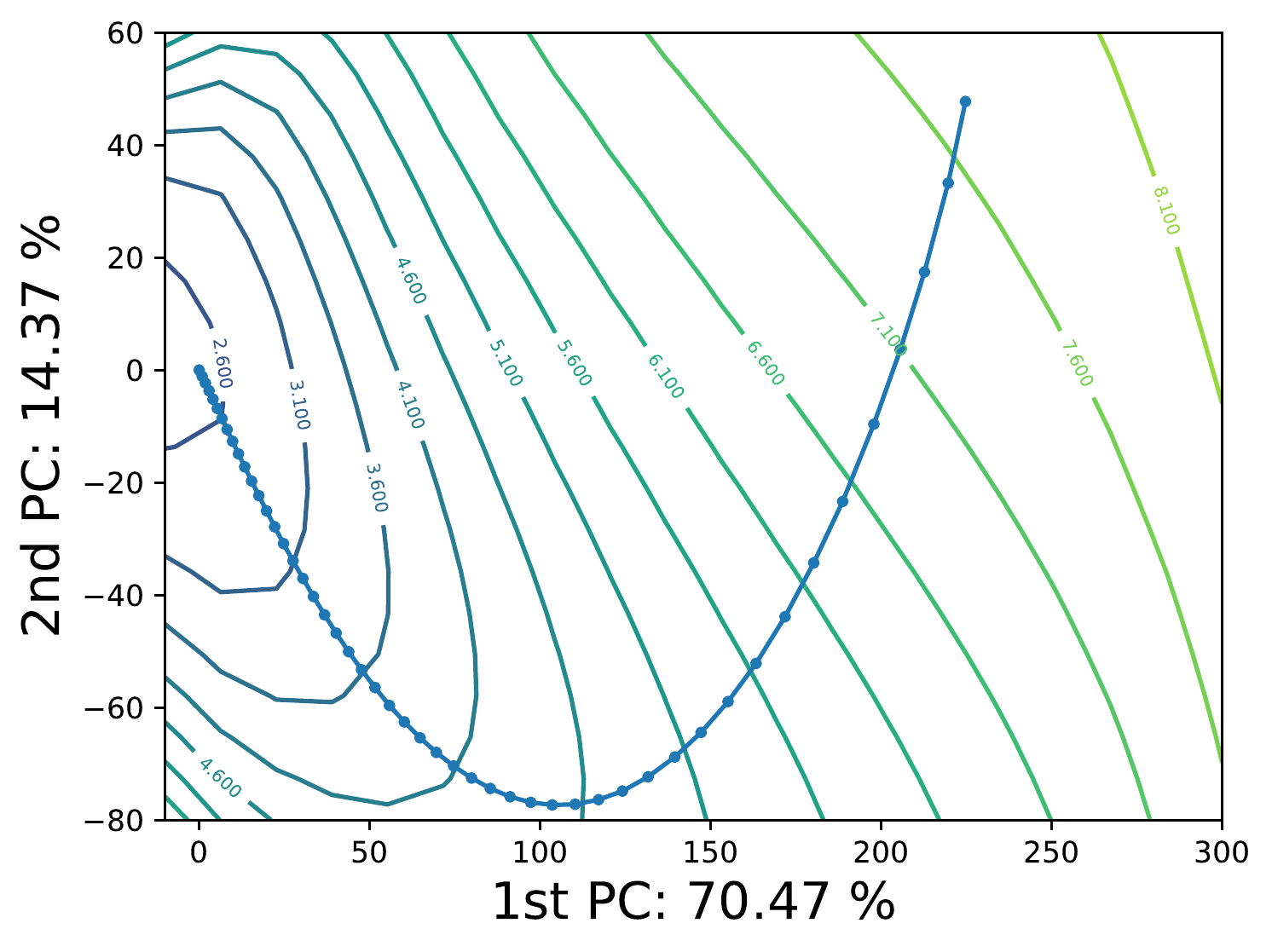}}               
\subfigure[adam-inv-5e-4]{\includegraphics[scale=0.35]{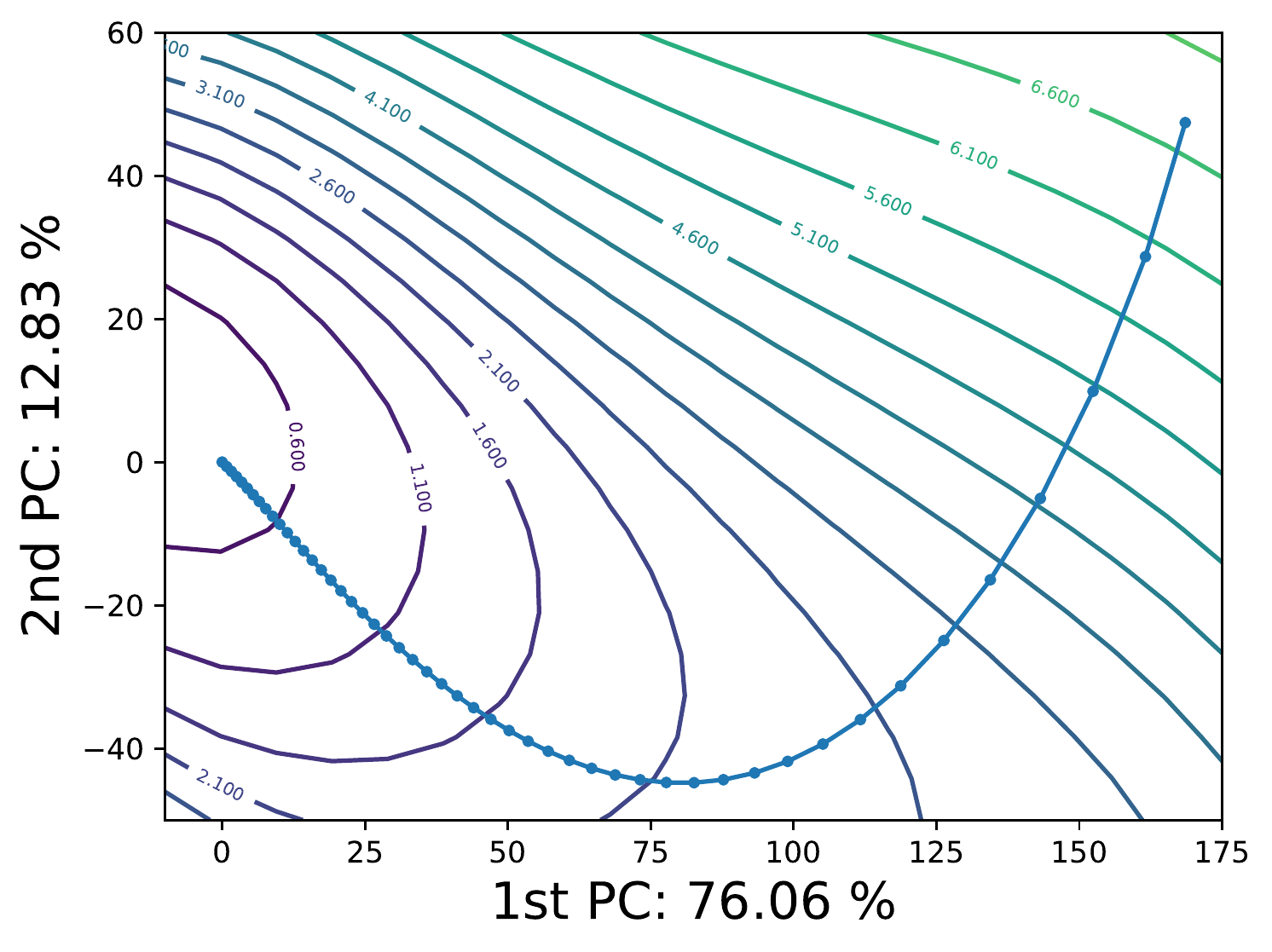}}                
\subfigure[adam-cyc-yshrink]{\includegraphics[scale=0.35]{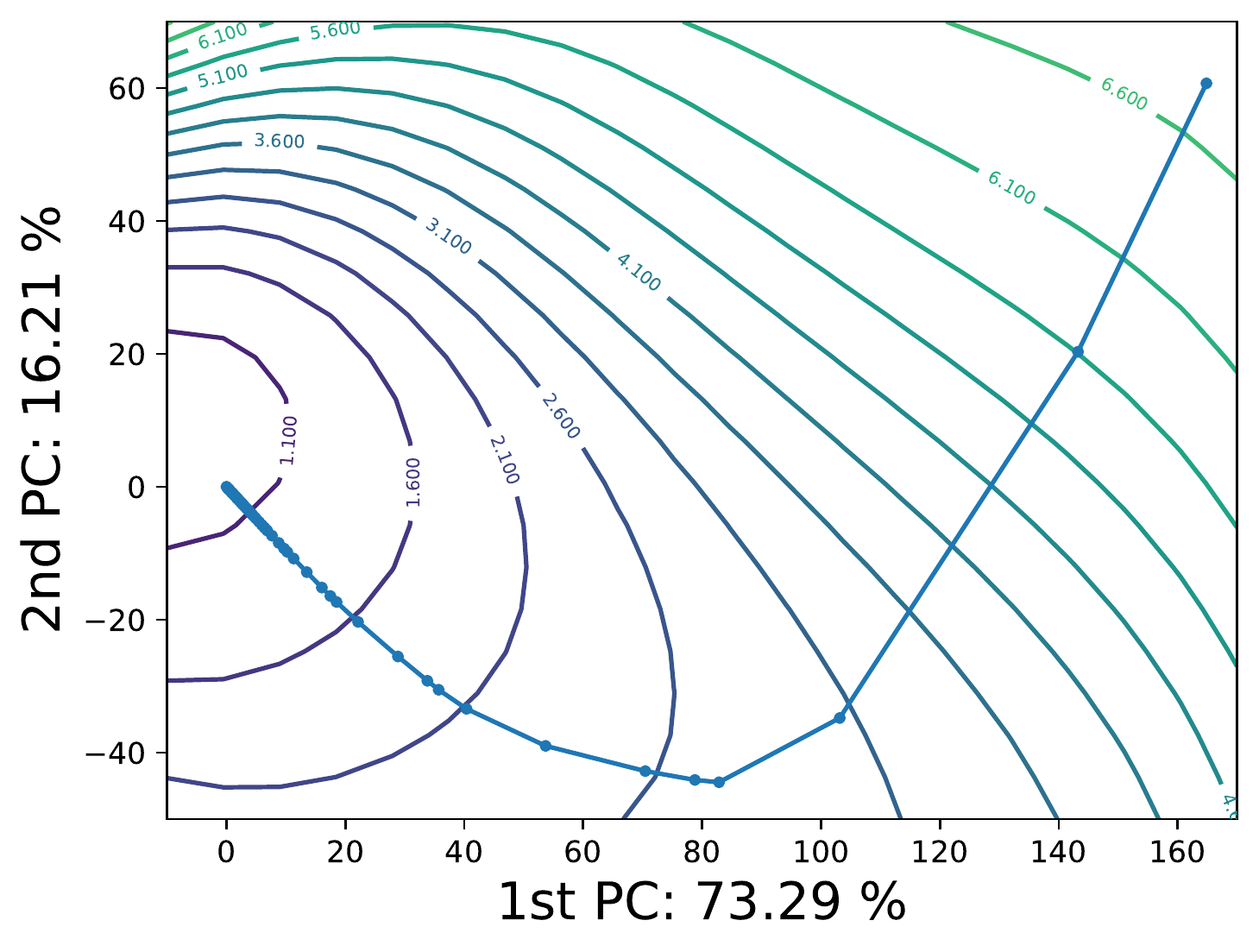}}                
\caption{Loss surface, optimizer trajectory and learning rates visualization for training NMT models on IWSLT2014-de-en.}                  
\label{fig:error_surface}                                                        
\end{figure*}

\subsection{Rationale behind Applying CLR to NMT Training}
There are two reasons proposed in \citet{clr} on why CLR works. The theoretical perspective proposed is that the increasing learning rate helps the optimizer to escape from saddle point plateaus. As pointed out in \citet{clr_saddle}, the difficulty in optimizing deep learning networks is due to saddle points, not local minima. The other more intuitive reason is that the learning rates covered in CLR are likely to include the optimal learning rate, which will be used throughout the training. Leveraging the visualization techniques proposed by \citet{viz_loss}, we take a peek at the error surface, optimizer trajectory and learning rate. The first thing to note is the smoothness of the error surface. This is perhaps not so surprising given the abundance of skip connections in transformer-based networks. Referring to Figure \ref{fig:error_surface} (c), we see the cyclical learning rate greatly amplifying Adam's learning rate in flatter region while nearer the local minimum, the cyclical learning rate policy does not harm convergence to the local minimum. This is in contrast to Figure \ref{fig:error_surface} (a) and (b), where although the adaptive nature of the learning rate in Adam helps to move quickly across flatter region, the effect is much less pronounced without the cyclical learning rate. Figure \ref{fig:error_surface} certainly does give credence to the hypothesis that the cyclical learning rate helps to escape saddle point plateaus, as well as the optimal learning rate will be included in the cyclical learning rate policy.

Some explanation about Figure \ref{fig:error_surface} is in order here. Following \citet{viz_loss}, we first assemble the network weight matrix by concatenating columns of network weights at each epoch. We then perform a Principal Component Analysis (PCA) and use the first two components for plotting the loss landscape. Even though all three plots in Figure \ref{fig:error_surface} seem to converge to the local minimum, bear in mind that this is only for the first two components, with the first two components contributing to 84.84\%, 88.89\% and 89.5\% of the variance respectively. With the first two components accounting for a large portion of the variance, it is thus reasonable to use Figure \ref{fig:error_surface} as a qualitative guide.


\section{Conclusion}
From the various experiment results, we have explored the use of CLR and demonstrated the benefits of CLR for transformer-based networks unequivocally. Not only does CLR help to improve the generalization capability in terms of test set results, but it also allows using larger batch size for training without adversely affecting the generalization capability. Instead of just blindly using default optimizers and learning rate policies, we hope to raise awareness in the NMT community the importance of choosing a useful optimizer and an associated learning rate policy.

\label{sect:conclusions}

\bibliography{CLR_NMT}
\bibliographystyle{acl_natbib}

\appendix

\section{Appendices}
\label{sec:appendix}
\begin{figure}[h]
    \centering
    \includegraphics[width=0.47\textwidth]{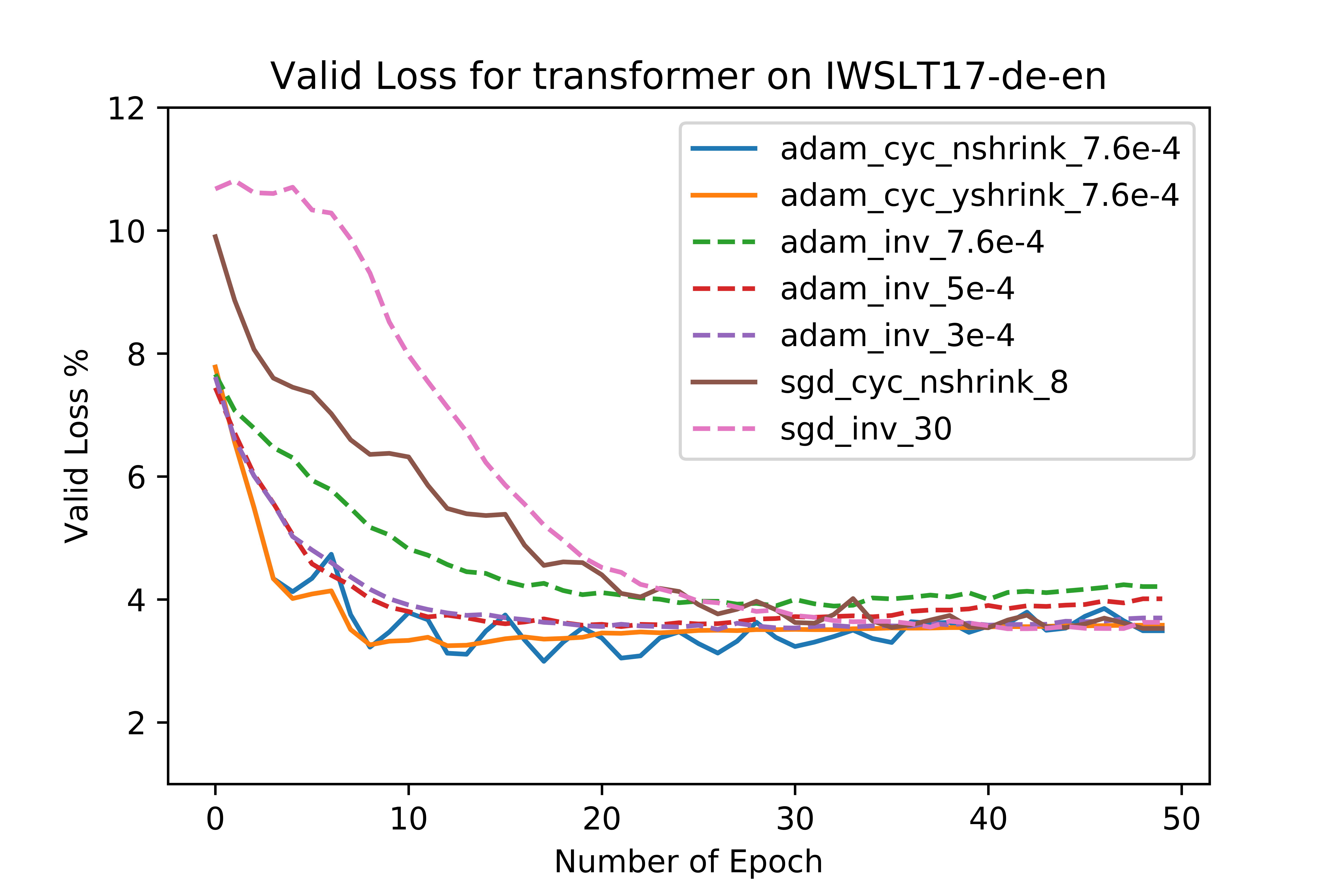}
    \caption{Effects of applying CLR to training NMT on IWSLT2017-de-en.}
    \label{figapendix1}
\end{figure}

\begin{figure}[h]
    \centering
    \includegraphics[width=0.47\textwidth]{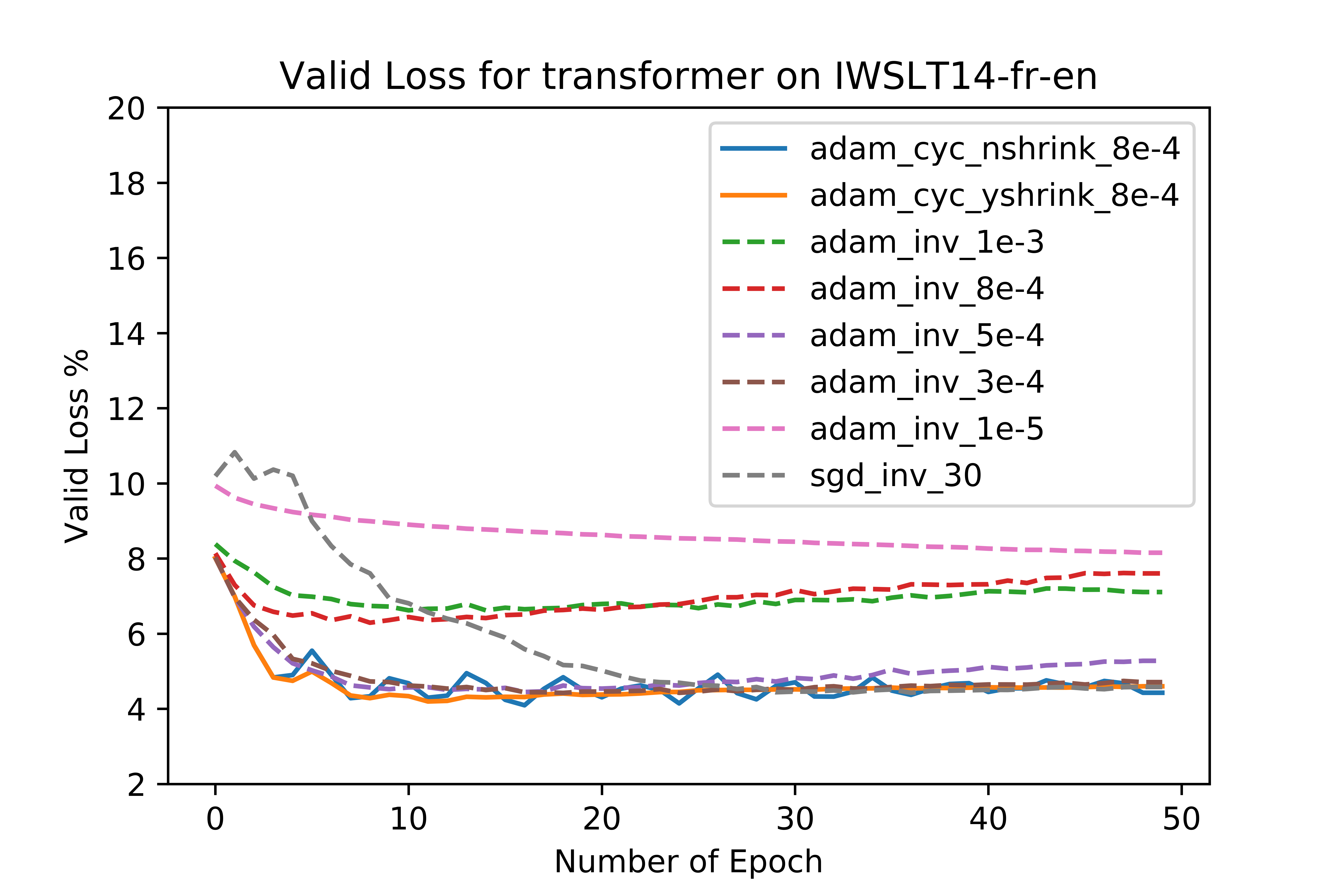}
    \caption{Effects of applying CLR to training NMT on IWSLT2014-fr-en.}
    \label{figappendix2}
\end{figure}

Figures~\ref{figapendix1}, ~\ref{figappendix2} are included in this Appendix.

\section{Supplemental Material}
\label{sec:supplemental}
Scripts and data are available at https://github.com/nlp-team/CL\_NMT.

\end{document}